\documentclass[runningheads]{llncs}
\usepackage{multirow}
\usepackage{hyperref,cite}
\usepackage[T1]{fontenc}
\usepackage{amsmath}
\usepackage{graphicx}
\usepackage[usenames,dvipsnames]{color}
\usepackage{amsfonts}
\begin{document}
\title{LesiOnTime - Joint Temporal and Clinical Modeling for Small Breast Lesion Segmentation in Longitudinal DCE-MRI}

\titlerunning{LesiOnTime}
\author{Mohammed Kamran \inst{1,2} 
\and Maria Bernathova \inst{3} 
\and Raoul Varga \inst{4}
\and Christian F. Singer \inst{5,7}
\and Zsuzsanna Bago-Horvath \inst{6,7}
\and Thomas Helbich \inst{4}
\and Georg Langs \inst{1,2}
\and Philipp Seeböck \inst{1,2}}
\authorrunning{M. Kamran et al.}
\institute{
     Computational Imaging Research Lab, Department of Biomedical Imaging and Image-guided Therapy, Medical University of Vienna, Austria \\ \email{mohammed.kamran, philipp.seeboeck@meduniwien.ac.at}
    \and Comprehensive Center for AI in Medicine, Medical University of Vienna, Austria
    \and Breast Health Center, Privatklinik Döbling, Austria 
    \and Department of Biomedical Imaging and Image-guided Therapy, Medical University of Vienna, Austria
    \and Department of Obstetrics and Gynecology, Medical University of Vienna, Austria
    \and Department of Pathology, Medical University of Vienna, Austria
    \and Comprehensive Cancer Center, Medical University of Vienna, Austria
    }
\maketitle              % typeset the header of the contribution
\begin{abstract}

Accurate segmentation of small lesions in Breast Dynamic Contrast-Enhanced MRI (DCE-MRI) is critical for early cancer detection, especially in high-risk patients. While recent deep learning methods have advanced lesion segmentation, they primarily target large lesions and neglect valuable longitudinal and clinical information routinely used by radiologists. In real-world screening, detecting subtle or emerging lesions requires radiologists to compare across timepoints and consider previous radiology assessments, such as the BI-RADS score. We propose \textit{LesiOnTime}, a novel 3D segmentation approach that mimics clinical diagnostic workflows by jointly leveraging longitudinal imaging and BI-RADS scores. The key components are: (1) a \textit{Temporal Prior Attention (TPA)} block that dynamically integrates information from previous and current scans; and (2) a \textit{BI-RADS Consistency Regularization (BCR)} loss that enforces latent space alignment for scans with similar radiological assessments, thus embedding domain knowledge into the training process. Evaluated on a curated in-house longitudinal dataset of high-risk patients with DCE-MRI, our approach outperforms state-of-the-art single-timepoint and longitudinal baselines by 5\% in terms of Dice. Ablation studies demonstrate that both TPA and BCR contribute complementary performance gains. These results highlight the importance of incorporating temporal and clinical context for reliable early lesion segmentation in real-world breast cancer screening. Our code is publicly available at \href{https://github.com/cirmuw/LesiOnTime}{https://github.com/cirmuw/LesiOnTime}

\end{abstract}

\section{Introduction}

Breast cancer is the most prevalent cancer among females, with about 2.3 million diagnosed cases and 670,000 deaths reported globally in 2022 alone~\cite{who2024breastcancer}. In the United States alone, an estimated number of 316,950 new invasive breast cancer cases are projected for 2025~\cite{acs_breastcancer}. Early detection significantly improves the 5-year survival rate, up to 90\%~\cite{witowski2022improving}.

2D Mammography-based screening is widely accepted as an effective technique for early detection\cite{kebede2024dual}. However, for individuals at elevated risk—such as those with a family history or pathogenic mutations conferring a lifetime risk of >30\%, 3D Dynamic Contrast-Enhanced MRI (DCE-MRI) is recommended due to its superior sensitivity \cite{lee2010breast}. Radiologists assess DCE-MRI using the Breast Imaging-Reporting and Data System (BI-RADS) (0–6) to estimate malignancy likelihood and guide follow-up, where scores $\leq$3 suggest benignity, and $\geq$4 indicate malignancy probabilities between 2\% and 95\%, warranting further workup. However, accurate detection of small or subtle lesions (median volume of 800 mm³ or even less) in DCE-MRI remains challenging due to varying size, shape, enhancement patterns and low contrast with adjacent tissues. Normal background parenchymal enhancement (BPE) can mimic or mask lesions, especially in premenopausal women. There is also significant intra- and inter-reader variability, even among experienced radiologists with years of training. In addition, inconsistent acquisition protocols and a lack of large, annotated datasets hinder the development of robust AI segmentation systems.

\paragraph{Related Work}
The availability of annotated breast DCE-MRI datasets has driven progress in automated lesion analysis~\cite{garrucho2025large,danielsstandard,zhao2023breastdm,saha2018machine}. While early work focused on 2D slice-wise tumor segmentation~\cite{carvalho2021tumor, zhang2018breast}, recent approaches leverage 3D models~\cite{garrucho2025large,zhao2023breastdm}, contrastive learning~\cite{guo2023contrastive}, and diffusion models~\cite{konz2024anatomically}. GAN-based anomaly detection has shown promise for identifying subtle early changes in high-risk women, even before visible lesions emerged~\cite{burger2023deep}. Along the same line of research, pre-diagnostic scans from high-risk cohorts up to one year before cancer identification have shown predictive potential~\cite{hirsch2025early}, highlighting the importance of longitudinal data. State-of-the-art segmentation models include CNN-based nnUNet \cite{isensee2021nnu} and transformer-based architectures such as SwinUNETR \cite{he2023swinunetr} and SAM \cite{kirillov2023segment}. Longitudinal extensions like \textit{LongiSeg} \cite{rokuss2024longitudinal} have been proposed in other domains (e.g., multiple sclerosis lesion segmentation in brain MRI), but adaptation to breast MRI remains largely unexplored. 
Moreover, most studies still target large lesions in non–high-risk populations \cite{garrucho2025large, zhao2023breastdm, zhao2024biomedparse} and rely on single time-point segmentation models \cite{he2023swinunetr, isensee2021nnu, kirillov2023segment}, neglecting longitudinal context from prior scans \cite{garrucho2025large, zhao2023breastdm, danielsstandard} or clinical information such as BI-RADS scores \cite{rokuss2024longitudinal}—data routinely available in clinical follow-ups. While \textit{LongiSeg} \cite{rokuss2024longitudinal} represents a step toward longitudinal segmentation, it requires the availability of manual pixel-wise annotations of previous timepoints, which are typically unavailable in breast screening scenarios where lesions are absent or undetectable at earlier timepoints. Moreover, it ignores complementary clinical context such as BI-RADS scores, limiting its utility in real-world diagnostic workflows.

\paragraph{Contribution} 
To address these limitations, we propose \textit{LesiOnTime}, a novel longitudinal segmentation approach for breast DCE-MRI. It integrates temporal imaging (DCE-MRI) and clinical information (BI-RADS scores) via two main innovations, the Temporal Prior Attention (TPA) Block and BI-RADS Consistency Regularizer (BCR). 
In contrast to \cite{rokuss2024longitudinal}, our approach does not require a manual pixel-wise lesion annotation for the previous timepoint, as the TPA block dynamically learns to weight the relevance of longitudinal scans. For instance, if a previous scan shows very little change compared to the current one, the network can down-weight it. Conversely, in cases where the temporal progression is meaningful (e.g., subtle enhancement appearing over time), the model learns to emphasize it. 
Furthermore, we incorporate clinical information during training using the BCR loss. It encourages feature similarity across timepoints only when BI-RADS scores are consistent, and allows divergence when scores indicate clinical progression—thus enforcing both temporal and clinical consistency. 

Experiments on a curated high-risk cohort show that LesiOnTime surpasses state-of-the-art methods by 5\% Dice, including both single-timepoint and longitudinal models (nnUNet~\cite{isensee2021nnu}, SwinUNETR~\cite{he2023swinunetr}, LongiSeg~\cite{rokuss2024longitudinal}). Ablation studies confirm the complementary value of TPA and BCR: removing either component reduces performance, while their combination yields the best results.

\section{Method}

\begin{figure}[t]
	\centering
	\includegraphics[width=1.0\textwidth]{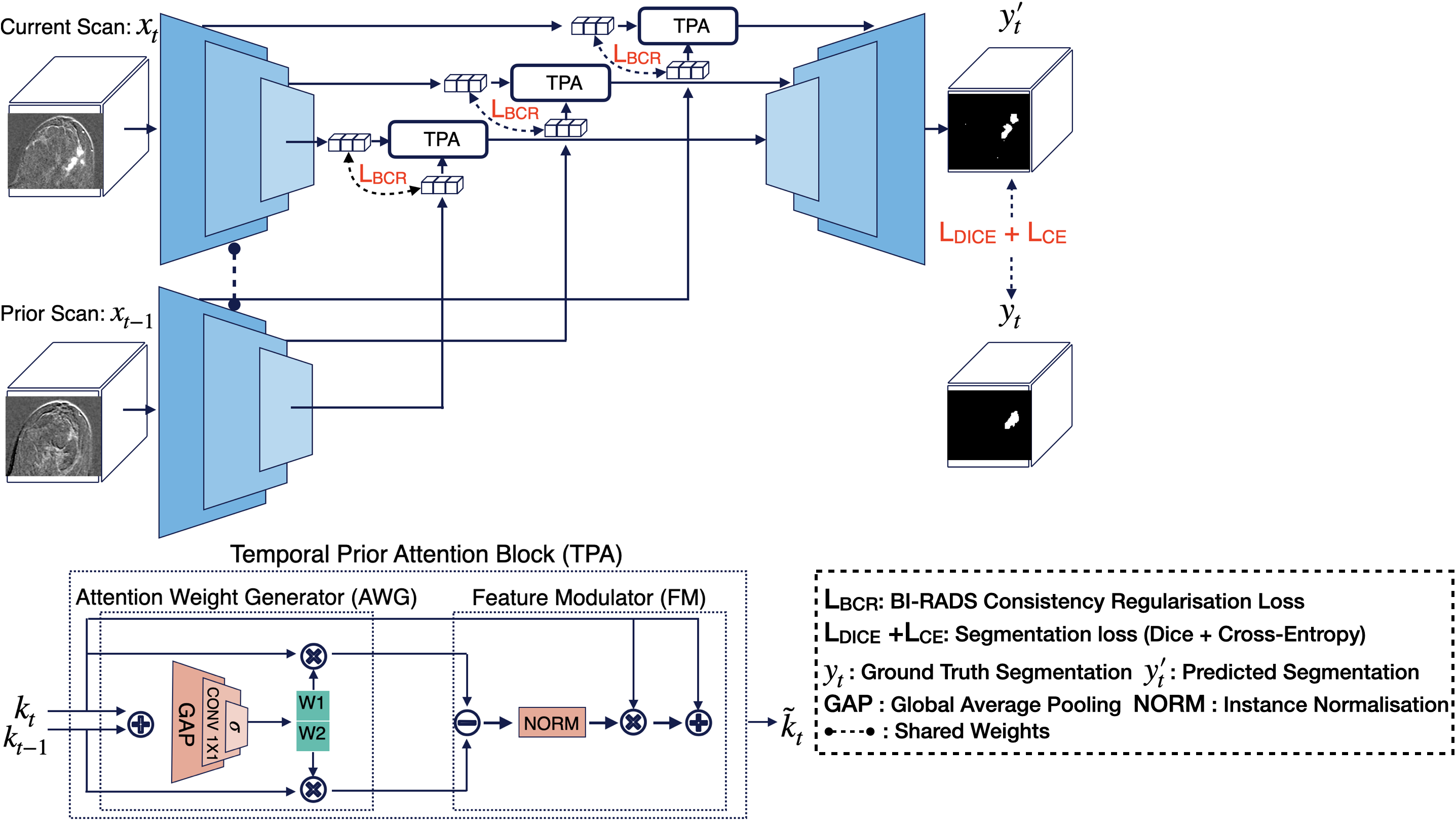}
	\caption{\textbf{LesiOnTime:} Paired current and prior scans ($x_{t}$,$x_{t-1}$) are processed by a dual-encoder with shared weights. Features from both encoders ($k_{t}$,$k_{t-1}$) pass through the proposed TPA block at each skip connection, which includes an attention weight generator (AWG) and a feature modulator (FM). AWG produces attention weights ($W_1$, $W_2$) for the scan pair, used by the FM to adaptively combine features. The BCR loss ($\mathcal{L}_{\text{BCR}}$) regularizes the latent space according to BI-RADS progression during training. The predicted segmentation map ($y'_t$) is compared with the ground truth ($y_t$) using a combination of Dice and Cross-Entropy loss ($\mathcal{L}_{\text{DICE}} + \mathcal{L}_{\text{CE}}$).
}
	\label{fig:method}
\end{figure}

We propose \textbf{LesiOnTime}, a novel 3D segmentation approach that jointly models temporal imaging features and clinical priors (Figure~\ref{fig:method}). It integrates temporal information through a temporal prior attention (TPA) block (Section~\ref{subsec:method_tpa}) and clinical context via a BI-RADS consistency regularization loss (BCR) (Section~\ref{subsec:method_bcr}).  
Formally, let $x_{t}, x_{t-1} \in \mathbb{R}^{C \times D\times H \times W}$ denote 3D breast DCE-MRI volumes for visits at the timepoints $t$ (current) and $t-1$ (previous), with $ y_{t} $ as the corresponding ground truth segmentation mask. The model $ f_\theta $, parameterized by $\theta$, takes the pair of volumes as input and predicts the segmentation mask of the current timepoint $t$, $y'_{t} = f_\theta(x_{t}, x_{t-1})$.

LesiOnTime adopts a shared-weight dual encoder-decoder architecture tailored for longitudinal lesion analysis. The model processes pair of scans $x_{t}, x_{t-1}$ (current and prior) through two parallel encoders with shared weights, ensuring consistent feature representations across visits. At each layer \( m \), the encoded feature maps from the current scan \( k_t^m \) and the prior scan \( k_{t-1}^m \) are passed into a \textbf{Temporal Prior Attention (TPA)} block, which captures and modulates temporal dependencies. 

The model is trained end-to-end using a composite loss: (1) the \textbf{segmentation loss} combining \textbf{Dice loss} and \textbf{Cross-Entropy loss} for accurate lesion delineation, and (2) the proposed \textbf{BI-RADS Consistency Regularization (BCR) loss} (Section~\ref{subsec:method_bcr}), which constrains feature embeddings in accordance with clinical progression.
\begin{equation}
\mathcal{L}_{\text{total}} = \lambda_{dice}\mathcal{L}_{\text{Dice}} + \lambda_{ce}\mathcal{L}_{\text{CE}} + \lambda_{bcr} \mathcal{L}_{\text{BCR}}
\end{equation}

%The TPA block includes an \textbf{Attention Weight Generator (AWG)}, consisting of feature concatenation, global average pooling (GAP), a 1×1 convolution, and a sigmoid activation (\( \sigma \)), which generates adaptive attention weights. These weights modulate the prior features before passing through a \textbf{Feature Modulator (FM)} that subtracts the modulated prior from the current features, normalizes the result, and reintegrates it with the current features. This process yields the temporally enhanced representation \( \tilde{k}_t^m \). 

\subsection{Temporal Prior Attention (TPA) Block}
\label{subsec:method_tpa}

The TPA block consists of two main components: the \textit{attention weight generator} and the \textit{feature modulator}. Given feature maps \( k_{t}^{m}, k_{t-1}^{m} \in \mathbb{R}^{c \times d \times h \times w} \) from the dual encoders at layer \( m \), they are concatenated along a new dimension into a new tensor of shape \( \mathbb{R}^{2 \times c \times d \times h \times w} \).

The AWG applies global average pooling across the spatial dimensions ($D$, $H$, $W$), reducing each channel to a scalar. The resulting vector is passed through a 1D convolution layer and a sigmoid activation to generate soft attention weights \( w = [w_1, w_2] \in \mathbb{R}^{2 \times 1} \), where \( w_1 \) and \( w_2 \) are scalar values representing the attention assigned to visits \( t \) and \( t-1 \), respectively.

The FM uses these learned attention weights to adaptively adjust the current feature map \( k_t^m \) in relation to the prior feature map \( k_{t-1}^m \). This is done through a weighted residual difference:
\begin{equation}
\tilde{k}_{t}^{m} = k_{t}^{m} \odot \operatorname{InstNorm}(w_1 \odot k_{t}^{m} - w_2 \odot k_{t-1}^{m}) + k_{t}^{m},
\end{equation}
where \( \odot \) denotes element-wise (broadcasted) multiplication across channels. This formulation allows the model to emphasize meaningful temporal changes while preserving the core features of the current scan.

By dynamically adjusting attention across each patient’s scan pair, the TPA block learns to model cross-scan interactions. This enables the network to focus on the most informative timepoint, enhancing sensitivity to subtle or slowly evolving lesions.

\subsection{BI-RADS Consistency Regularizer}
\label{subsec:method_bcr}
The BI-RADS Consistency Regularization (BCR) loss is designed to enforce \textbf{clinical and temporal coherence} in feature learning across longitudinal breast MRI scans. Specifically, it constrains the distance between feature embeddings of consecutive scans based on the change in BI-RADS scores. 
At each layer $m$, the feature-level BCR loss is defined as:
\begin{equation}
\mathcal{L}^{(m)}_{\text{feat}} = \frac{\tanh\left( \left\| k_{t}^{(m)} - k_{t-1}^{(m)} \right\|^2 \right)}{\Delta \text{BI-RADS} + \epsilon},
\end{equation}
%with the limiting behaviour,
%\begin{equation}
%\lim_{\substack{\Delta \text{BI-RADS} \to 0 \\ \left\| k_t^{(m)} - k_{t-1}^{(m)} \right\|^2 \to 0}} \mathcal{L}_{\text{feat}}^{(m)} = 0
%\end{equation}
where $\Delta \text{BI-RADS}$ is the absolute difference in BI-RADS scores between visits $t$ and $t-1$, and $\epsilon$ is a small positive constant to avoid division by zero. The \textit{tanh} activation bounds feature differences in the numerator to [0,1), reducing sensitivity to extreme feature differences and promoting stable training. 
The overall BCR loss across a set of all layers $\mathcal{M}$ is computed as a weighted sum:
\begin{equation}
\mathcal{L}_{\text{BCR}} = \sum_{m \in \mathcal{M}} w_m \cdot \mathcal{L}^{(m)}_{\text{feat}},
\end{equation}
with $w_m$ denoting layer-wise weights. Clinically, a stable BI-RADS score implies minimal pathological change, motivating feature similarity across timepoints. Conversely, an increase in BI-RADS typically reflects the emergence of new enhancement or growth of lesions in the current scan relative to prior imaging, promoting greater feature divergence. The BCR loss explicitly encodes this behavior, as it approaches zero when the BI-RADS scores are identical and the embeddings are well aligned, enabling joint modeling of longitudinal imaging and radiological assessment.

%%%%%%%%%%%%%%%%%%%%%%%%%%%%%%%%%%%%%%%%%%%%%%%%%%%%%%%%%%%%%%%%%%%%%%%%%%%%%%%%%%%%%%%%%%%%%%%%%%%%%%%%%%%%%%%%
%%%%%%%%%%%%%%%%%%%%%%%%%%%%%%%%%%%%%%%%%%%%%%%%%%%%%%%%%%%%%%%%%%%%%%%%%%%%%%%%%%%%%%%%%%%%%%%%%%%%%%%%%%%%%%%%
\section{Experimental Setup}
\paragraph{Material} 
We utilize a curated in-house breast DCE-MRI dataset comprising 155 studies (BI-RADS 4/5) acquired at the Vienna General Hospital (AKH), each with pre-contrast and at least three post-contrast T1-weighted volumes with spatial resolutions of 48x384×384 or 80x512×512 and a median lesion volume of 800 mm³. Manual pixel-wise lesion annotations were created by an expert radiologist for all 155 visits, using ITK-SNAP on the first post-contrast volume. Each study has a prior scan 6–24 months apart, enabling longitudinal modeling (310 visits total). BI-RADS scores are available for all visits, generated by radiologists in clinical routine. Data are randomly split by patient into development (n=131, 5-fold cross-validation) and an independent test set (n=24) for final evaluation.

%\begin{table}[h]
%\centering
%\begin{tabular}{l c c c}
%\hline
%\textbf{Subset} & \textbf{Train} & \textbf{Val} & \textbf{Test} \\
%\hline
%Complete Set & 163 & 31 & 28 \\
%Longitudinal Subset & 110 & 21 & 24 \\
%\hline
%\end{tabular}
%\caption{Dataset counts for Train, Validation, and Test splits of Complete and Longitudinal subsets}
%\label{tab:dataset_counts}
%\end{table}

\paragraph{Preprocessing} 
All volumes are resampled to a 0.7 mm × 0.7 mm × 2 mm voxel spacing. We perform rigid and nonlinear registration of the current and previous timepoint volume using the Symmetric Normalization (SyN) algorithm of Advanced Normalization Tools Python (ANTsPy). 
Subsequently, we crop the volumes to the breast tissue area using a breast segmentation algorithm~\cite{perschy2023breast}, excluding thoracic structures that may confound lesion segmentation.

\paragraph{Experiments} 
We compare \textit{LesiOnTime} to LongiSeg \cite{rokuss2024longitudinal}, nnUNet \cite{isensee2021nnu}, SwinUNETR \cite{he2023swinunetr} (trained on 3D volumes), and BiomedParse \cite{zhao2024biomedparse}, a large 2D vision-language segmentation model (fine-tuned on 2D slices with prompts). All models are evaluated consistently in 3D, with the entire volume treated as single sample. 
Ablation experiments include \textit{LesiOnTime\textsubscript{w/o TPA}} (TPA replaced by difference-weighting of LongiSeg) and \textit{LesiOnTime\textsubscript{w/o BCR}} (BCR loss removed).

\paragraph{Training Details} 
All models are trained on an NVIDIA RTX 6000 GPU (48GB VRAM) using a batch size of 2 for 1000 epochs. Loss combines Dice, Cross-Entropy ($\lambda_\text{dice} = \lambda_\text{ce} = 1.0$), and BCR terms ($\lambda_\text{fc} = 0.1$). BCR weights $w_m$ are linearly scaled across network layers (0.1 to 1.0) with {$\epsilon$=1e-1}, emphasizing deeper, semantically richer features (e.g. 0.1,0.3,0.5,0.7,0.9,1 for a 6-layer network). Our longitudinal approach builds on the previously proposed architectural selection strategy to identify the most suitable U-Net configuration~\cite{rokuss2024longitudinal,isensee2021nnu}. The final architecture is a 3D PlainConvUNet from the dynamic network architectures library, with 6 stages in the encoder progressively increasing feature channels [32, 64, 128, 256, 320, 320] using 3D convolutions (Conv3d) and instance normalization with a kernel size of 3x3x3. The decoder mirrors this structure with 2 convolutional layers per stage. LeakyReLU is used as the nonlinearity, and the architecture avoids dropout.

%%%%%%%%%%%%%%%%%%%%%%%%%%%%%%%%%%%%%%%%%%%%%%%%%%%%%%%%%%%%%%%%%%%%%%%%%%%%%%%%%%%%%%%%%%%%%%%%%%%%%%%%%%%%%%%%
%%%%%%%%%%%%%%%%%%%%%%%%%%%%%%%%%%%%%%%%%%%%%%%%%%%%%%%%%%%%%%%%%%%%%%%%%%%%%%%%%%%%%%%%%%%%%%%%%%%%%%%%%%%%%%%%
\section{Results and Discussion}
\label{sec:results_discussion}
Table~\ref{tab:seg_results} summarizes segmentation performance on the held-out test set. \textit{LesiOnTime} achieves the highest Dice score (0.356), outperforming both longitudinal (LongiSeg: 0.302) and single-timepoint models (nnUNet: 0.291, SwinUNETR: 0.195, BiomedParse: 0.122). LesiOnTime also yields superior precision, recall, and HD-95 distance, indicating improved boundary delineation and lesion localization. Statistical comparison using the Wilcoxon signed-rank test revealed significant improvements in HD-95 over all baselines (p < 0.05), while differences in Dice score did not reach statistical significance. We hypothesize that this may be due to the known high variability of Dice scores for small structures, where boundary-focused metrics such as HD-95 are considered more stable and sensitive to subtle segmentation improvements~\cite{maier2024metrics}.
%
%\Pc{A Wilcoxon signed-rank test confirms that improvements over LongiSeg and nnUNet are statistically significant in terms of HD-95 ($p < 0.05$), underscoring the benefit of incorporating temporal and clinical priors.}

%nnUNET \cite{isensee2021nnu} consistently outperforms other non-longitudinal baselines. The longitudinal models consistently outperform the non-longitudinal ones, with \textit{LesiOnTime} achieving the highest Dice score (0.356). This demonstrates that incorporating temporal and patient-specific progression data helps the models better understand tumor boundaries.

\begin{table}[t]
\centering
\caption{Quantitative results of 3D breast lesion segmentation on the DCE-MRI test set. Best results are in bold. HD-95: 95th percentile Hausdorff distance.}
\begin{tabular}{lcccc}
\hline
 \textbf{Model} & \textbf{Dice} & \textbf{HD-95} & \textbf{Precision} & \textbf{Recall} \\
\hline
  nnUNET\cite{isensee2021nnu}      & 0.29 & 164.9 & 0.32 & 0.34 \\
  SwinUNETR \cite{he2023swinunetr}       & 0.195 & 182.3 & 0.24 & 0.25 \\
  BiomedParse \cite{zhao2024biomedparse}      & 0.12 & 192.2 & 0.14 & 0.20 \\
  LongiSeg~\cite{rokuss2024longitudinal}         & 0.30 & 132.2 & 0.31 & 0.34 \\
  \textbf{LesiOnTime}$_\mathrm{w/o~TPA}$(ours)         & 0.31  & 110.4 & 0.37 & 0.41 \\
  \textbf{LesiOnTime}$_\mathrm{w/o~BCR}$(ours)    & 0.32 & 112.3 & 0.38 & 0.37 \\
\textbf{LesiOnTime} (ours) & \textbf{0.35} & \textbf{106.5} & \textbf{0.39} & \textbf{0.42} \\
\hline
\label{tab:seg_results}
\end{tabular}
\end{table}

\begin{figure}[t]
	\centering
	\includegraphics[width=1.0\textwidth]{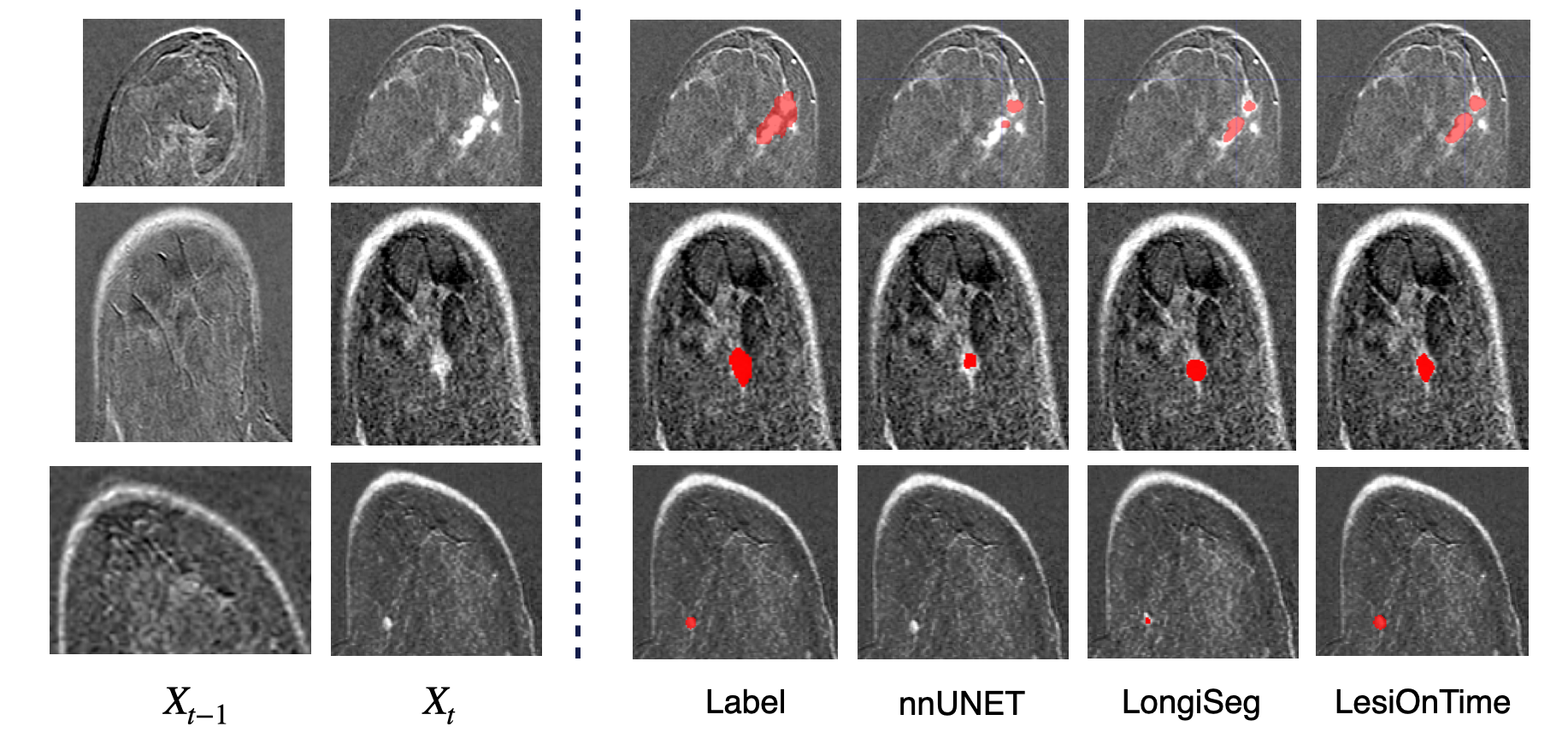}
	\caption{Qualitative segmentation results comparing \textit{LesiOnTime} with state-of-the-art baselines: single timepoint nnUNet~\cite{isensee2021nnu} and multi-timepoint LongiSeg~\cite{rokuss2024longitudinal}.}
	\label{fig:qualitative_results}
\end{figure}

This is in line with qualitative results (Figure~\ref{fig:qualitative_results}), where \textit{LesiOnTime} shows superior delineation of lesions, particularly for extremely small lesions and non-mass enhancements with diffuse boundaries — a persistent challenge due to low contrast and annotation ambiguity. We observed that \textit{LesiOnTime} seems to focus on areas of highest intensity and structural consistency, while radiologists may intentionally annotate larger regions to prioritize high sensitivity in ambiguous cases.

%We highlight two key observations. In the first row, the case illustrates scattered non-mass enhancement lesions, which are particularly difficult to segment. Expert annotations often encompass these regions as a single connected area due to time constraints or clinical prioritization, leading to coarse boundaries. Our method, however, learns to separate and localize these fine-grained lesion clusters more precisely, maintaining their non-mass-like structure.
%In other cases, radiologists may annotate larger regions to ensure high sensitivity, especially when lesion boundaries are ambiguous. While these annotations are clinically valid, our model is able to refine the segmentation by focusing on areas of highest intensity and structural consistency, thus improving boundary accuracy while remaining faithful to the expert labels.

%Our proposed method builds upon LongiSeg \cite{rokuss2024longitudinal} by replacing the conventional difference weighting block with a \textbf{Temporal Prior Attention (TPA)} incorporating temporal context and a \textbf{BI-RADS Consistency Regularizer Loss (BCR)}, integrating clinical radiology score.

\begin{figure}[t]
	\centering
	\includegraphics[width=0.4\textwidth]{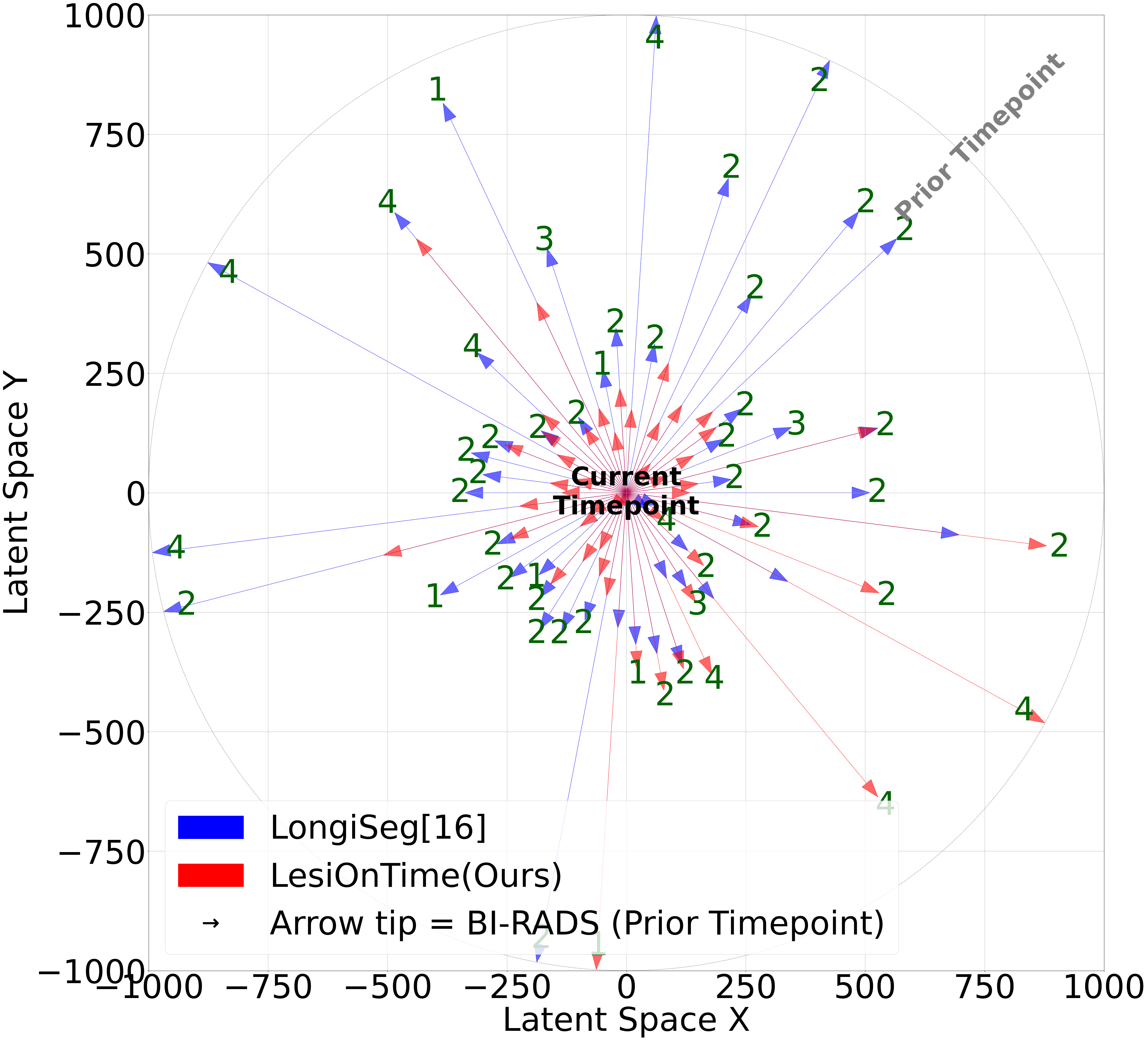}
	\caption{This radial plot illustrates vector embeddings of current (center) and prior (outer circle) timepoints in latent space, with arrow magnitudes representing the Euclidean distance between these embeddings. It compares the effect of training with BCR (\textbf{LesiOnTime}, red) versus without BCR (LongiSeg, blue). Arrow tips are labeled with BI-RADS scores, indicating clinical suspicion at the prior scan, while all current timepoints have a fixed BI-RADS score of 4.}
	\label{fig:results_embedding}
\end{figure}

{Figure~\ref{fig:results_embedding} shows that BCR guides latent features to cluster according to BI-RADS scores. Specifically, distances between current and prior timepoints shrink for the majority of cases when trained with BCR (LesiOnTime), indicating better alignment in latent space. Without BCR (LongiSeg\cite{rokuss2024longitudinal}), these distances are generally larger, underscoring BCR’s role in embedding clinically relevant priors.}

One limitation of this study is its single-center setting. Currently, no publicly available longitudinal breast DCE-MRI dataset exists for external validation. However, the method's utilization of temporal and radiological priors supports transferability to other sites and scanners. Notably, \textit{LesiOnTime} requires no manual lesion annotation for prior timepoints and uses BI-RADS scores only during training, not inference, simplifying clinical deployment. 
While segmentation of small, diffuse lesions remains challenging (e.g. due to low uptake contrast or scattered nature), leveraging prior timepoints partially mitigates this, as shown by improved segmentation over single-timepoint baselines. This highlights the advantage of incorporating temporal context for lesion detection in screening populations. 
We excluded public datasets such as BreastDM~\cite{zhao2023breastdm}, MAMA-MIA~\cite{garrucho2025large}, DUKE MRI~\cite{saha2018machine} and Advanced-MRI-Breast-Lesions~\cite{danielsstandard} due to their focus on large or late-stage lesions and the lack of longitudinal scans, unrepresentative of our clinical scenario focused on subtle early-stage lesions. 
Although registration accuracy was visually validated in a subset of cases, minor misalignments cannot be fully excluded and may contribute to residual segmentation errors. However, the TPA mechanism was designed to reduce sensitivity to such discrepancies by weighting temporal features adaptively. 

%Moreover, segmentation performance on extremely small and diffuse lesions remains challenging - especially due to their sparse and scattered nature / non-mass enhancements with weak contrast uptake, however... the use of multi-timepoint data is promising as demonstrated with results in this paper
%Accurate pixel-wise segmentation of small lesions in breast MRI is highly challenging, especially due to their sparse and scattered nature. Since only a few pixels are annotated, even small misclassifications can significantly reduce the Dice score. On the other hand, majority of Breast DCE-MRI public datasets \cite{zhao2023breastdm,danielsstandard,garrucho2025large,saha2018machine} work on large lesions. As a result, we excluded them from our evaluation study. 

%%%%%%%%%%%%%%%%%%%%%%%%%%%%%%%%%%%%%%%%%%%%%%%%%%%%%%%%%%%%%%%%%%%%%%%%%%%%%%%%%%%%%%%%%%%%%%%%%%%%%%%%%%%%%%%%
%%%%%%%%%%%%%%%%%%%%%%%%%%%%%%%%%%%%%%%%%%%%%%%%%%%%%%%%%%%%%%%%%%%%%%%%%%%%%%%%%%%%%%%%%%%%%%%%%%%%%%%%%%%%%%%%
\section{Conclusion}
In this work, we present \textit{LesiOnTime}, a novel breast DCE-MRI segmentation approach that explicitly integrates longitudinal imaging context and radiological assessments to address key limitations of conventional single-timepoint models. Through its temporal prior attention (TPA) mechanism, \textit{LesiOnTime} leverages prior scans without requiring manual annotations for earlier timepoints. The proposed BI-RADS Consistency Regularization (BCR) aligns latent representations with diagnostic labels, embedding clinical reasoning into the model. \textit{LesiOnTime} consistently outperforms both longitudinal and single-timepoint baselines in segmenting small, diffuse, and clinically challenging lesions typical for high-risk screening populations. These findings highlight the value of temporal and clinical cues for improving early breast cancer segmentation in MRI, and suggest that \textit{LesiOnTime} may support future clinical workflows by enabling more reliable and context-aware lesion assessment in longitudinal breast cancer screening.

\subsubsection*{Acknowledgements}
This work was supported by the European Federation for Cancer Images (EUCAIM, Grant No. 101100633), Initiative Krebsforschung – Grant 2022 (UE77104006), and Vienna Science and Technology Fund (WWTF, PREDICTOME) [10.47379/LS20065]. %The authors gratefully acknowledge the support provided by these funding bodies.

\subsubsection*{Disclosure of Interests}
GL: meduniwien.ac.at, mit.edu, csh.ac.at, contextflow.com; PS: meduniwien.ac.at; imperial.ac.uk; univie.ac.at; TH: imperial.ac.uk; bayer.com; bd.com; novomed.at; guerbet.com; siemens.com; hologic.de; bruker.com; ZBH: MSD, Astra Zeneca, Gilead, Stemline, Abbvie.

%In this work, we propose \textbf{LesiOnTime}, a 3D breast DCE-MRI segmentation framework that integrates temporal priors and clinical knowledge. By leveraging prior scans via temporal prior attention (TPA) and aligning latent features with radiological assessments through BI-RADS-constrained regularization (BCR), LesiOnTime overcomes key limitations of single-timepoint methods. It surpasses state-of-the-art baselines in segmenting subtle and challenging lesions typical of high-risk screening populations. Our design requires no manual annotations for prior scans and relies on BI-RADS labels only during training—enhancing practicality for clinical translation. These findings underline the importance of temporal and clinical cues for improving early breast cancer detection in MRI.

%\subsubsection{Acknowledgements} Please place your acknowledgments at
%the end of the paper, preceded by an unnumbered run-in heading (i.e.
%3rd-level heading).

%\newpage
\bibliographystyle{splncs04} % LNCS style

%\bibliography{bibliography} % replace "bibliography" with the name of your .bib file (without the .bib extension)
\end{document}